\documentclass[a4paper,10pt,conference]{ieeeconf}

\IEEEoverridecommandlockouts
\usepackage{copyright}

\usepackage[english]{babel}

\usepackage{amssymb}
\usepackage{amsmath}
\usepackage{hyperref}
\usepackage{cleveref}
\usepackage{graphicx}
\usepackage{latexsym}
\usepackage{algorithm}
\usepackage{algpseudocode}
\usepackage{newclude}
\usepackage{multicol}
\usepackage{tabularx}
\usepackage{booktabs}
\usepackage{threeparttable}
\let\labelindent\relax
\usepackage{enumitem}
\usepackage{listings}

\usepackage[labelformat=simple]{subcaption}

\usepackage{combelow}
\usepackage{cite}

\usepackage[export]{adjustbox}

\usepackage{url}
\newcommand\rurl[1]{%
  \href{http://#1}{\nolinkurl{#1}}%
}

\crefname{table}{Table}{Tables}
\crefname{figure}{Figure}{Figures}
\crefname{section}{Section}{Sections}

\usepackage[binary-units]{siunitx}
\usepackage{xcolor}

\usepackage[nonumberlist]{glossaries}
\newacronym{wsl}{WSL}{Weakly-Supervised Learning}
\newacronym{fmcw}{FMCW}{Frequency-Modulated Continuous Wave}
\newacronym{ro}{RO}{Radar Odometry}
\newacronym{vo}{VO}{Visual Odometry}
\newacronym{ml}{ML}{Machine Learning}
\newacronym{dl}{DL}{Deep Learning}
\newacronym{cnn}{CNN}{Convolutional Neural Network}
\newacronym{svm}{SVM}{Support Vector Machine}
\newacronym{gps}{GPS}{Global Positioning System}
\newacronym{dft}{DFT}{Discrete Fourier Transform}
\newacronym{stft}{STFT}{Short Time Fourier Transform}
\newacronym{fft}{FFT}{Fast Fourier Transform}
\newacronym{gmm}{GMM}{Gaussian Mixture Model}
\newacronym{mfcc}{MFCC}{Mel Frequency Cepstral Coefficient}
\newacronym{dct}{DCT}{Discrete Cosine Transform}
\newacronym{slam}{SLAM}{Simultaneous Localisation and Mapping}
\newacronym{hmm}{HMM}{Hidden Markov Model}
\newacronym{fcn}{FCN}{Fully Convolutional Network}
\newacronym{ekf}{EKF}{Extended Kalman Filter}
\newacronym{rnn}{RNN}{Recurrent Neural Network}
\newacronym{slic}{SLIC}{Simple Linear Iterative Clustering}
\newacronym{lidar}{LiDAR}{Light Detection and Ranging}
\newacronym{tof}{TOF}{time-of-flight}
\newacronym{osm}{OSM}{OpenStreetMap}

\glsdisablehyper

\renewcommand{\baselinestretch}{0.97}

\begin{document}

%------------------------------------------------------------------
\title{\bf Keep off the Grass: Permissible Driving Routes\\from Radar with Weak Audio Supervision}
\author{David Williams$^*$, Daniele De Martini$^*$, Matthew Gadd$^*$, Letizia Marchegiani$^{\dagger}$, and Paul Newman$^*$
\\
$^*$Oxford Robotics Institute, Dept. Engineering Science, University of Oxford, UK.\\\texttt{\{dw,daniele,mattgadd,pnewman\}@robots.ox.ac.uk}
\\
$^{\dagger}$Automation and Control, Dept. Electronic Systems, Aalborg University, DK.\\\texttt{lm@es.aau.dk}
\thanks{This project is supported by the Assuring Autonomy International Programme, a partnership between Lloyd’s Register Foundation and the University of York, as well as UK EPSRC Programme Grant EP/M019918/1.
Additionally, we would like to thank the Groundskeepers and Officers of the University Parks as well as our partners at Navtech Radar.}
}
\maketitle
%------------------------------------------------------------------

\copyrightnotice

%------------------------------------------------------------------
\begin{abstract}
Reliable outdoor deployment of mobile robots requires the robust identification of permissible driving routes in a given environment.
The performance of \acrshort{lidar} and vision-based perception systems deteriorates significantly if certain environmental factors are present e.g. rain, fog, darkness.
Perception systems based on \acrlong{fmcw} scanning radar maintain full performance regardless of environmental conditions and with a longer range than alternative sensors.
Learning to segment a radar scan based on driveability in a fully supervised manner is not feasible as labelling each radar scan on a bin-by-bin basis is both difficult and time-consuming to do by hand. 
We therefore weakly supervise the training of the radar-based classifier through an audio-based classifier that is able to predict the terrain type underneath the robot. 
By combining odometry, \acrshort{gps} and the terrain labels from the audio classifier, we are able to construct a terrain labelled trajectory of the robot in the environment which is then used to label the radar scans.
Using a curriculum learning procedure, we then train a radar segmentation network to generalise beyond the initial labelling and to detect all permissible driving routes in the environment.
\end{abstract}

\glsresetall
\glsunset{lidar}
\glsunset{gps}

%------------------------------------------------------------------
\section{Introduction}%
\label{sec:introduction}
%------------------------------------------------------------------

Safe navigation of intelligent mobile robots in unstructured and unknown outdoor environments (e.g. search and rescue, agriculture, and mining industry sectors) requires perception systems which deliver a detailed understanding of surroundings regardless of any environmental factor (e.g. weather, scene illumination, etc). 
% As each terrain type is uniquely challenging to traverse, robust route identification is a key problem to be solved.
In many environments, some terrains are unsuitable to traverse and so robust route identification is a key problem to be solved.
% To that end, a variety of sensor technologies can be used for terrain identification, including: cameras, \gls{lidar}, sonar, audio, and radar. 
To that end, a variety of sensor technologies have been used for solving related problems, including: cameras, \gls{lidar}, sonar, audio, and radar. 

\gls{lidar}- and vision-based terrain classification systems are highly susceptible to inclement environmental or atmospheric conditions: heavy rain, fog, direct sunlight, and dust all greatly degrade the performance of these systems, thereby limiting their range of applications.
% deep-learning systems based on vision susceptible to adversarial attacks

\gls{fmcw} scanning radar, in contrast, operates robustly under such adverse conditions and additionally operates at ranges of up to many hundreds of metres -- relaxing the maximum speed at which a robot can safely travel and facilitating longer planning horizons.
Indeed, there is a burgeoning interest in exploiting \gls{fmcw} radar to enable robust mobile autonomy, including ego-motion estimation~\cite{cen2018precise,aldera2019,2019ITSC_aldera,Barnes2019MaskingByMoving,UnderTheRadarArXiv}, localisation~\cite{KidnappedRadarArXiv,gadd2020lookaroundyou,tang2020rsl,UnderTheRadarArXiv}, and scene understanding~\cite{williams2019listening,weston2019probably,kaul2020rssnet}.

\begin{figure}
    \centering
    \includegraphics[width=0.8\columnwidth]{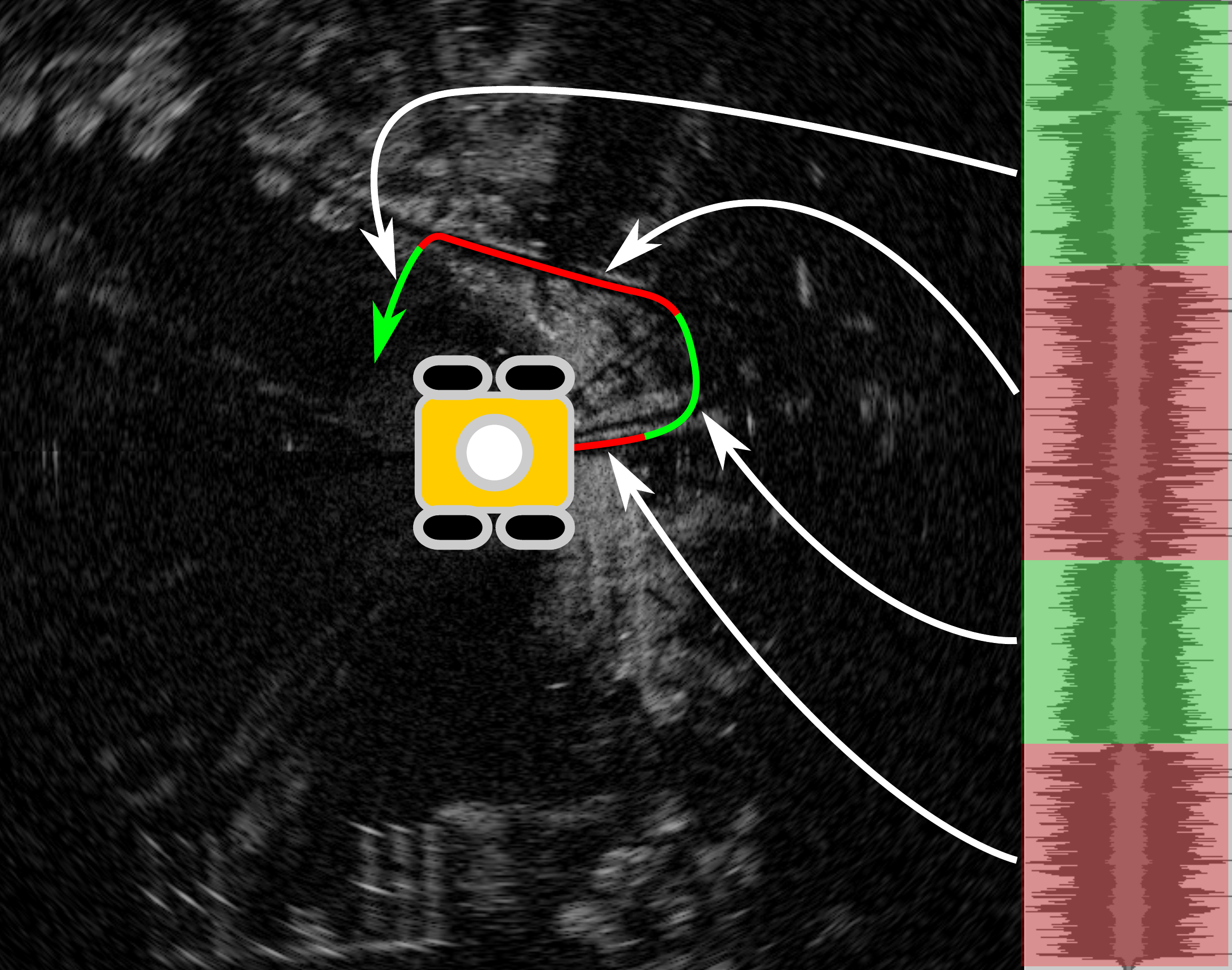}
    \caption{Overview of the proposed system: audio is recorded and used to classify the terrain the robot is driving on -- here gravel (red) and grass (green).
    Using odometry, the robot can paint this semantic information on top of the radar scan.
    \label{fig:sys_overview}}
    \vspace{-.65cm}
\end{figure}

As a novel contribution to scene understanding with radar, this paper presents a system that detects permissible driving routes from raw radar scans.
Specifically, it focusses on the methodology for the obtainment of labelling and a novel training procedure for the radar classifier.
% \cref{fig:sys_overview} illustrates aspects of the proposed system.

Radar measurements are complex, containing significant multipath reflections, speckle noise, and other artefacts in addition to the radar's internal noise characteristics~\cite{robo_radar}.
This makes the interaction of the electromagnetic wave in the environment more complex than that of \gls{tof} lasers.
% Thus, obtaining a labelled radar dataset for supervision -- each scan annotated on a bin-by-bin basis -- is challenging and time consuming.
% We therefore propose an weakly-supervised framework using an alternative sensing modality: audio.
As obtaining a labelled radar dataset for supervision -- with each scan annotated on a bin-by-bin basis -- is challenging and time consuming, we propose an weakly-supervised framework using an alternative sensing modality: audio.
%
% Audio-based terrain classification of these terrain types is made possible by the consideration that permissible driving routes are characterised fully by their terrain type (e.g. gravel, grass, asphalt) and by the fact that each interaction between the robot and the ground has a terrain-specific audio signature.

\begin{figure*}
    \centering

    \begin{subfigure}{0.3\textwidth}
    \includegraphics[width=\textwidth]{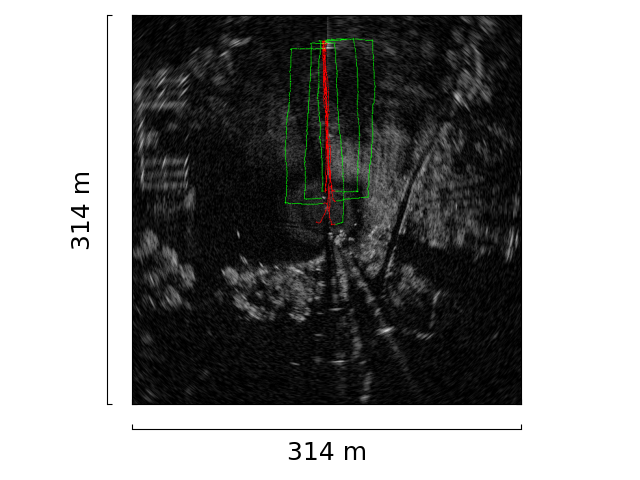}
    \caption{\label{fig:initial_labels}}
    \end{subfigure}
    %
    % {\LARGE$\xrightarrow{}$}
    % \quad
    %
    \begin{subfigure}{0.3\textwidth}
    \includegraphics[width=\textwidth]{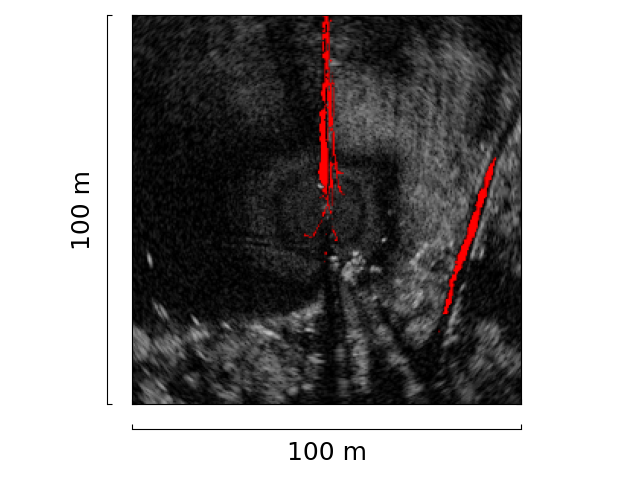}
    \caption{\label{fig:gen_labels}}
    \end{subfigure}
    %
    % {\LARGE$\xrightarrow{}$}
    % \quad
    %
    \begin{subfigure}{0.3\textwidth}
    \includegraphics[width=\textwidth]{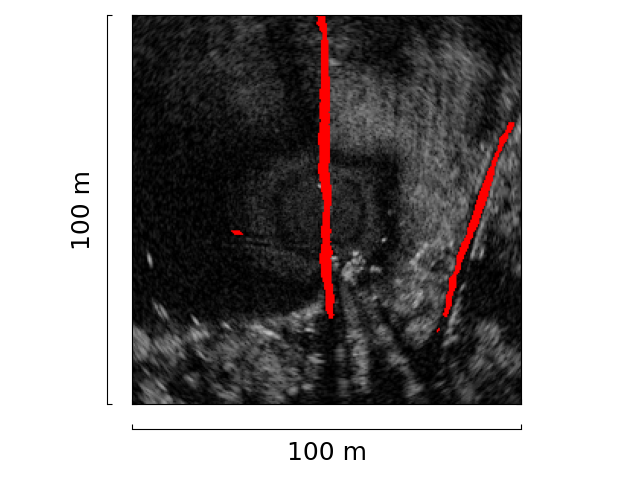}
    \caption{\label{fig:final_seg}}
    \end{subfigure}
    \caption{An example from the training dataset at three stages of the training process. (a) shows initial labelling (b) shows additional labels generated by stage one of the curriculum and (c) shows the final segmentation result.\label{fig:frontpage}}
    \vspace{-.6cm}
\end{figure*}

Audio-based terrain classifiers can be used to predict the permissibility of a driving route when the route is characterised by its terrain (e.g. grass, gravel, asphalt).
Predicting terrain from audio is possible as each interaction between the robot and the ground has a terrain-specific audio signature.

Audio offers two advantages over other modalities, e.g.~vision-based systems: first, audio is invariant to scene appearance and less affected by weather conditions, providing more stable and predictable results; moreover, the use of microphones is advantageous as audio is a one-dimensional signal, easing the labelling process as the audio for each terrain can be collected separately.

Once the audio-based terrain classifier has been trained, we exploit it to weakly supervise the radar classifier training.
\gls{vo} and \gls{gps} are used to trace the trajectory of the robot on the radar scan as if it were a canvas (see \cref{fig:sys_overview}) and each traversed bin is classified by the audio classifier.
% In theory, with access to \gls{gps}, it should be possible extract labels for the audio from \gls{osm}.
% Thus, the system could be trained in a completely self-supervised fashion.
% We leave this to future work.

% This paper proceeds by reviewing existing literature in~\cref{sec:related}.
% \Cref{sec:method,sec:application} present the theoretical and practical details for the implementation.
% \cref{sec:results} reports our results.
% \cref{sec:conclusions} discusses the contribution and suggests future avenues of investigation.

%------------------------------------------------------------------
\section{Related Work}%
\label{sec:related}
%------------------------------------------------------------------

Mature techniques for identifying the driveable area of urban environments with cameras and \glspl{lidar} often learn to semantically segment the entire scene through the use of fully labelled datasets such as Cityscapes~\cite{cordts2016cityscapes} or by weak supervision and demonstration as in~\cite{barnes16}.
In non-urban outdoor environments, path detection is closely related to the task of terrain classification~\cite{blas2008fast}.
For the environment in which our system was trained and tested, all permissible driving routes belong to one terrain class (gravel) and so for this application the tasks of permissible driving route identification and terrain classification are equivalent.

Vision-based terrain classification is perhaps the most traditional approach due to its associated intuitiveness and affordability.
In~\cite{jansen2005colour}, colour segmentation is employed to identify different terrains, while~\cite{blas2008fast} performs both colour and texture segmentation for path detection.
However in~\cite{jansen2005colour}, problems arising due to variations in illumination are exposed.
% In addition to this, sources of artificial illumination such as headlights are also an inexact substitution for natural light, causing yet more complication. 
Although these problems are mitigable, when also paired with environmental factors such as fog, heavy rain and dust clouds, these systems alone seem unfit for robust autonomy.

\gls{lidar} can be used to build successful terrain classifiers by observing the texture of the 3D point-cloud as seen in~\cite{kragh2015object}.
In low light conditions \gls{lidar} works well, however it suffers greatly in the presence of rain and fog, limiting its applicability in much the same way as vision.

%%%%%% AUDIO %%%%%%

As mentioned in \cref{sec:introduction}, audio can also be used for terrain classification.
Terrain-specific audio signatures are invariant to scene appearance and much less influenced by weather conditions compared with vision and \gls{lidar}-based methods.
The obvious disadvantage to this technique is that only the terrain the robot is currently operating on can be classified. 
As discussed in this paper, this characteristic can be leveraged for labelling purposes.
\cite{Valada2018} reports classification of nine different terrains with an accuracy of \SI{99.41}{\percent} by leveraging advances in \gls{dl} and using a \gls{cnn} classifier. 
The audio features used for the \gls{cnn} classifier were spectrograms generated with the \gls{stft}.

%%%%%% RADAR %%%%%%

Despite the lower spatial resolution and compression of height information, in~\cite{mielle2019comparative} it is shown in the context of a \gls{slam} system that while producing slightly less accurate maps than \glspl{lidar}, radars are capable of capturing details such as corners and small walls.
This is reflected in literature as extensive research has been done using millimetre-wave radar systems for odometry, obstacle detection, mapping and outdoor reconstruction~\cite{cen2018precise,heuer2014detection,robo_radar}.
Radar is invariant to almost all environmental factors posed by even the most extreme environments, such as dusty underground mines, blizzards~\cite{brooker2007seeing,foessel1999short}.
Less work, however, has been carried out to investigate radar's performance on more comprehensive scene understanding tasks such as terrain classification or path identification.
\cite{reina2011radar} presents an outdoor ground segmentation technique using a millimetre wave radar, however the chosen method limits its range of operation.
%The technique relies on the existence of the \textit{ground echo}, which is a high intensity parabola seen in a radar scan where the electromagnetic wave interacts with the ground.

Perhaps most similar to our work is a visual terrain classifier which is also supervised by learned acoustic features presented in~\cite{zurn2019self}.
In our work, however, we focus on the usage of radar, which has advantages over vision in terms of robustness to both weather and illumination as well as sensor range.
This work exposes at the same time challenges specific to the modality -- especially the high sparsity of labelling.
This is overcome with a stronger focus on the training procedure for the proposed network by explicitly promoting generalisation.

\begin{figure*}
  \centering
  \vspace{-.5cm}
  \begin{subfigure}[b]{0.27\linewidth}
    \includegraphics[width=\linewidth]{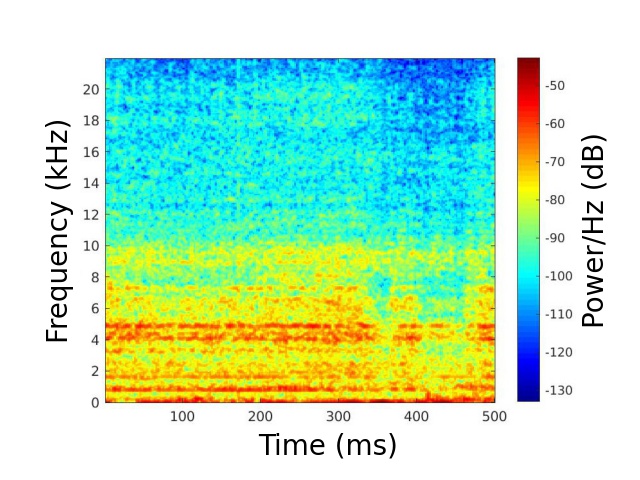}
     \caption{Spectrogram\label{fig:spectogram}}
  \end{subfigure}
  \begin{subfigure}[b]{0.27\linewidth}
    \includegraphics[width=\linewidth]{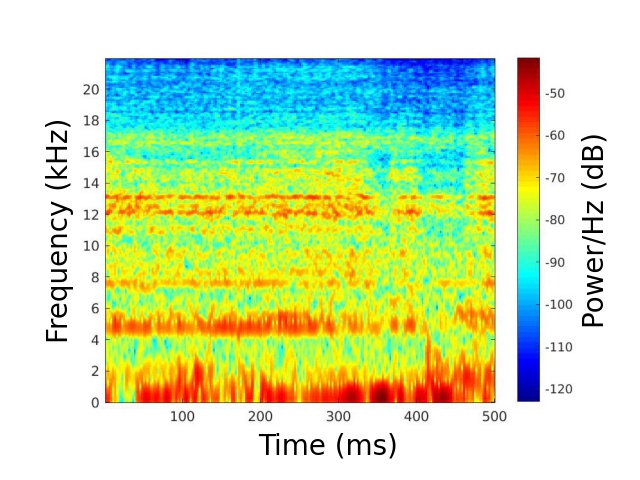}
    \caption{Mel-frequency spectrogram\label{fig:mel_frequency}}
  \end{subfigure}
  \begin{subfigure}[b]{0.27\linewidth}
    \includegraphics[width=\linewidth]{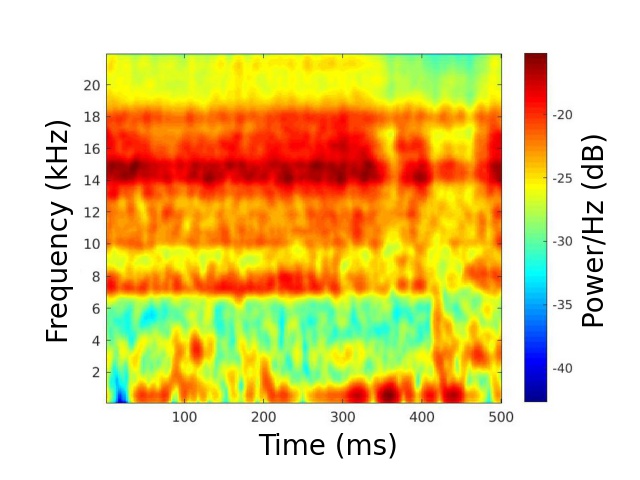}
    \caption{Gammatonegram\label{fig:gammatonegram}}
  \end{subfigure}
  \caption{Visualisation of each time-frequency diagram used as feature representation for audio. Each diagram is generated from a clip length of \SI{0.5}{\second}, shows frequencies up to half the sampling frequency and uses a bandwidth resolution of \SI{100}{\hertz}.\label{fig:tf1}}
  \vspace{-.6cm}
\end{figure*}

%------------------------------------------------------------------
\section{Methodology}%
\label{sec:method}

Our method is based on our early investigation described in~\cite{williams2019listening}.
Learning to segment driveable routes in a radar scan -- in a supervised manner -- requires that routes in each scan are labelled.
For a dataset of sufficient size (in the order of thousands of training examples), doing this by hand is a prohibitively time-consuming process.
We therefore opt to weakly supervise the training of a radar-based segmentation network with an audio-based classifier that is trained independently of the radar-based classifier.
Audio is collected for each terrain separately (making labelling trivial by tagging each sample as the whole sequence) and used to train the audio classifier for later use. 
Although the same labelling strategy could have been used to label radar directly, using audio as the labelling signal greatly increases the flexibility of the system by allowing multiple passes on different terrains in the same sequence and thus adding examples of both classes on single frames and at the same time lowering the sparsity of the labels.
%- The paper motivates the labelling through audio data. How were the handlabels for the audio generated? Couldnt the same procesdure be used to directly label the radar in a supervised manner, what is the advantage of the audio supervision? \ddm{...}

Through the use of odometry and \gls{gps}, we obtain the data collection robot's timestamped trajectory in the environment.
The audio terrain classifier is then used to accurately predict the terrain at each timestamp.
By combining both, we produce a terrain-labelled trajectory of the robot in the environment (depicted in \cref{fig:sys_overview}) which is used as sparse labelling.
%Using an \gls{ekf}, the labelled trajectories are calibrated and overlaid on each scan to produce a sparse labelling.
For the purpose of segmenting paths in our environment, only the terrain labels denoting gravel are required.

\subsection{Audio Classification}

As audio is best interpreted as a sequence of frequencies correlated in time, we discuss its representation in the form of different types of spectrograms.
As suggested in \cite{Valada2018}, spectrograms can be used as 1-channel images to feed into a \gls{cnn}.
This is effective as the success of \gls{cnn} classifiers is in their ability to learn features automatically from data containing local spatial correlations.
By assuming local spatial correlations in a spectrogram, the classifier recognises the temporal correlation of characteristic audio frequencies.

% The \gls{cnn} comprises of several convolutional layers followed by max-pooling layers and concludes with fully-connected and softmax activation layers to output the terrain class.

Our \gls{cnn} classifier follows a standard architecture with several convolutional layers and max-pooling for downsampling. 

For audio representation, we assess the performance of three types of spectrograms (results found in \cref{sec:results}).
The representations considered are: Spectrograms, Mel-frequency spectrograms and Gammatonegrams (see~\cref{fig:tf1}). 

Spectrograms are the simplest time-frequency diagrams and are generated directly by the \gls{stft}.
Mel-frequency spectrograms and gammatonegrams are motivated by the idea that the human auditory system does not perceive pitch in a linear manner. 
For humans, lower frequencies are perceptually much more important than higher frequencies and this can be represented in time-frequency representations.
Gammatonegrams extend this biological inspiration, using filter banks modelled on the human cochlea and have been successfully used before in a robotics context \cite{Marchegiani2018}.

The implementation used to generate both spectrograms and mel-frequency spectrograms is courtesy of VOICEBOX: Speech Processing Toolbox for MATLAB\footnote{Found at \url{www.ee.ic.ac.uk/hp/staff/dmb/voicebox/voicebox.html}. Produced by Mike Brookes, Dept. Electrical and Electronic Engineering, Imperial College in 1997.} and the MATLAB toolbox: Gammatone-like spectrograms\footnote{Found at \url{www.ee.columbia.edu/~dpwe/LabROSA/matlab/gammatonegram/}. Produced by Dan Ellis, Dept. of Electrical Engineering, Columbia University in 2009.} is used to generate gammatonegrams.

\subsection{From Audio to Labelled Radar}
In order to project terrain labels from audio into radar scans, we make use of the visual odometry estimate on the platform and \gls{gps}.
\gls{vo} produces a locally accurate, smooth trajectory and contains important orientation estimates.
Although the estimates are locally accurate, they tend to drift over longer distances.
In contrast, \gls{gps} measurements are globally accurate, but suffer from significant noise resulting in a non-smooth trajectory and contain low quality information about the orientation of the robot.
In order to leverage the benefits of both techniques, we fuse these data streams using an \gls{ekf}.

Once the robot's trajectory has been generated, it is labelled using the audio classifier to predict the terrain for each timestamp.
Finally, the labelled trajectory is fitted automatically to each radar scan using the position and orientation estimates from the \gls{ekf}.
% Once the trajectory is hand calibrated to the first scan, the position and orientation of the trajectory for the next scan can be calculated using the \gls{ekf} estimates.

\subsection{Radar Segmentation Training Procedure}
The nature of the method used for collecting the labels means that the radar scans are both inexactly and sparsely labelled.
The inexactness comes from measurement errors from the \gls{gps} and \gls{vo}, and the sparsity comes from our inability to thoroughly traverse every driveable surface observed in the radar scans.
This means that the training procedure must be designed such that the network can learn a more complex model than the labelling might immediately suggest. 

To do this, data augmentation and a label propagation technique are used to design a two stage curriculum learning procedure.
As described in~\cite{curriculum_learning}, the idea of curriculum learning is that neural networks perform better when presented with the most understandable training examples first.
This is done in the first stage by limiting the network's receptive field by only showing the network very small crops of the global scan.
In this way, the network is restricted to simply learning what a path looks like and is relieved of learning more complex concepts such as scene context. 
By comparison to the more difficult task of simultaneously segmenting multiple paths in the global scan, the network generalises much better on the simpler task of segmenting small crops (as suggested in~\cite{curriculum_learning}).
For this reason, we are able to generalise beyond the initially incomplete labelling (see \cref{fig:frontpage}).
Before input to the network, crops are also flipped, rotated, elastically deformed and rescaled to expose the network to paths that are of different orientations, shapes and widths.
This data augmentation promotes a broader understanding of how a path looks, thus assisting with generalisation.

Upon completion of the first stage, the network accurately segments small sections of paths contained in crops of the global scan (whether initially labelled or not) but is unsuited to segmenting the whole scan. 
The second stage of the curriculum is therefore to train the network to segment a whole scan containing multiple paths in one forward pass.
By combining the predictions of the network from stage one and the original labelling, we obtain a more complete and exact set of labels from which the network can be trained to complete the more complex task.
The idea of using a trained network's predictions to augment the labels is presented in a classification context in~\cite{labelrefinery}, however we adapt it to a segmentation context (described in \cref{sec:results:radar}).
% To generate the additional labels, the radar scan is divided into a small sub-scans which are sequentially segmented by the network.
% The new labels are generated by dividing each radar scan into a grid and segmenting each grid tile individually using the network from stage one.

For the segmentation network, we chose a U-Net architecture~\cite{ronneberger2015u}, which has proven effective for segmentation of radar scans~\cite{aldera2019,weston2019probably}.
A U-Net is a \gls{fcn} containing downsampling and upsampling paths with skip connections between paths to propagate fine detail.
%------------------------------------------------------------------
\section{Experimental Setup}%
\label{sec:application}
%------------------------------------------------------------------

This section discusses the platform and the dataset collected and used for training and testing of our system.

\subsection{Platform and Sensors}

A Clearpath Husky A200 robot was fitted with microphones and radar, for audio recording and route identification, and with cameras and \gls{gps} for odometry estimation.
The audio data was recorded by using two Knowles omnidirectional boom microphones, mounted in proximity to the two front wheels, and an ALESIS IO4 audio interface, at a sampling frequency of \SI{44.1}{\kilo\hertz} and a resolution of 16 bits.

We used a Navtech CTS350-X \gls{fmcw} scanning radar, mounted on top of the platform with an axis of rotation perpendicular to the road.
The radar operates at a frequency of \SIrange{76}{77}{\giga\hertz}, yielding  up to \SI{3600}{} range readings, each constituting one of the \SI{400}{} azimuth readings with a scan rotation rate of \SI{4}{\hertz}.
% In short range configuration, the range resolution is \SI{0.0438}{\metre} and in long range configuration it is \SI{~0.1752}{\metre} relating to a range of \SI{~157}{\metre} and \SI{630}{\metre} respectively. 
The radar's range resolution in short and long range configurations is \SI{0.0438}{\metre} and \SI{~0.1752}{\metre} respectively, resulting in ranges of \SI{~157}{\metre} and \SI{630}{\metre}.

Images for \gls{vo} were gathered by a Point Grey Bumblebee 2 camera, mounted facing the direction of motion on the front of the platform.
%The camera is characterised by \num{1024 x 768 x 3} resolution, \SI{20}{\hertz} FPS, 
%\num{1/3}$\inchsign$ Sony ICX204 CCD, global shutter, \SI{3.8}{\milli\metre} lens, \SI{66}{\degree} HFoV, \SI{12/24}{\centi\metre} baseline.
\gls{gps} measurements were collected with a GlobalSat BU-353-S4 USB GPS Receiver.

\subsection{Dataset}
As discussed in \cref{sec:method}, audio was collected for each terrain separately.
It was recorded from both microphones for \SI{15}{\minute} per terrain class, corresponding to approximately 7200 spectrograms per class (using a clip length of \SI{0.5}{\second}).
Audio for grass and gravel terrains was collected in University Parks and the asphalt terrain in the Radcliffe Observatory Quarter.

% For training and testing the radar classifier, two separate datasets were collected in two different locations in University Parks, Oxford.
Datasets for training and testing the classifier were collected with the radar in both the long range and short range configurations to ensure the network performs well regardless of specific radar configuration. 
We collected training data in two locations in University Parks, Oxford and testing data in two different locations in the same park. 
% In both datasets, the robot traverses both gravel and grass and labelling is provided by the audio classifier for training dataset.
The audio classifier in combination with \gls{vo} and \gls{gps} provides labelling for the training datasets. 
\cref{fig:frontpage} shows one location where the training dataset was collected comprises of two paths surrounded by grass.
% Due to the long range nature of the radar, it is unpractical to traverse every terrain or path observed by the radar.
% 98596 m2 for short range
As the radar scan covers an area of \SI{1587600}{\metre\squared} in its longest range configuration, it is impractical to traverse every path observed by the radar.
For this reason, we leave the side path untraversed (and therefore unlabelled), such that we can test the segmentation network's ability to generalise effectively.

%As previously mentioned, radar is characteristically difficult to densely label and so we are unable to calculate segmentation accuracies on the test dataset. In place of this, we present and discuss the test results qualitatively in an exhaustive fashion. 

% ------------------------------------------------------------------
\section{Results}%
\label{sec:results}
%------------------------------------------------------------------

This section presents experimental evidence of the efficacy of our system.

\subsection{Reliability of the Audio Supervisory Signal}

An investigation was performed into the performance of the audio classifier using each different audio feature representation to determine which one would be used in the final classifier.
In our experiments, the classifier is tested on a withheld testing dataset and predicts from three possible terrains: grass, gravel and asphalt.
After averaging over multiple experiments, the accuracies for the spectrogram, mel-frequency spectrogram and gammatonegram were \SI{98.5}{\percent}, \SI{98.8}{\percent}, \SI{99.4}{\percent} respectively (using a clip length of \SI{0.5}{\second}).
As the best performing feature representation, the gammatonegram was used to train the final audio terrain classifier.

Additionally, investigations into the audio clip length used to generate the gammatonegrams showed that the longer the clip length, the more accurate the terrain classifier.
Whilst an intuitive result, this means a compromise between accuracy and system frequency is necessary.
We chose a clip length of \SI{0.5}{\second} by balancing classification accuracy and other system frequencies (such as GPS update rate at \SI{1}{\hertz}) to result in a classification frequency of \SI{2}{\hertz}.

Lastly, the final audio terrain classifier was tested on a dataset where the robot dynamically traversed gravel and grass for \SI{22}{\minute}.
Approximate hand-labels were generated by cross-referencing the predicted terrain at each of the 1320 GPS measurements with satellite imagery. 
Here, the audio terrain classifier performed the task with an accuracy of \SI{98.4}{\percent}.

\subsection{Effective Supervision of Radar-only Segmentation}
\label{sec:results:radar}

Firstly, a U-Net is trained on the training set shown in~\cref{fig:initial_labels} as stage one in the curriculum detailed in~\cref{sec:method}.
Trained on the simple task of segmenting \num{64}$\times$\num{64} crops out of a \num{512}$\times$\num{512} scan, the network effectively learns not only to reproduce the labelling but also to segment paths unlabelled in our datasets (see \cref{fig:gen_labels}).

To generate the labels for the previously unlabelled sections of scans, the radar scan is divided into a small sub-scans which are sequentially segmented by the trained network.
To suppress spurious predictions, we randomly rotate each scan a small number of times and combine the predictions on each.
\cref{fig:frontpage} shows an example of both the initial labelling and the generated labelling after stage 1.

\begin{figure*}
    \centering
    \begin{subfigure}{0.22\textwidth}
    \includegraphics[width=0.8\textwidth, cframe=blue 2pt]{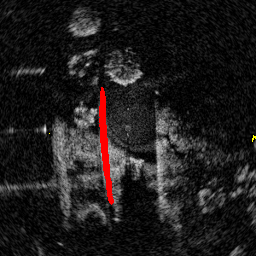}
    \caption{}
    \label{fig:1562077104209647}
    \end{subfigure}
    \begin{subfigure}{0.22\textwidth}
    \includegraphics[width=0.8\textwidth, cframe=blue 2pt]{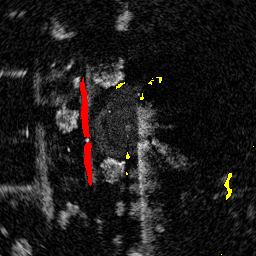}
    \caption{}
    \label{fig:1562077019044260}
    \end{subfigure}
    \begin{subfigure}{0.22\textwidth}
    \includegraphics[width=0.8\textwidth, cframe=cyan 2pt]{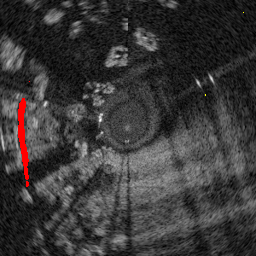}
    \caption{}
    \label{fig:1545308602535826}
    \end{subfigure}
    \begin{subfigure}{0.22\textwidth}
    \includegraphics[width=0.8\textwidth, cframe=cyan 2pt]{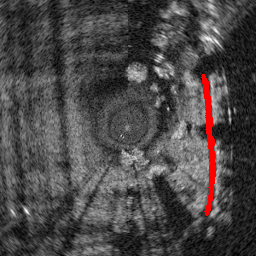}
    \caption{}
    \label{fig:1545308439330558}
    \end{subfigure}
    
    \begin{subfigure}{0.22\textwidth}
    \includegraphics[width=0.8\textwidth, cframe=blue 2pt]{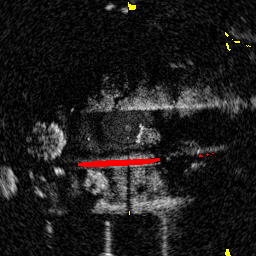}
    \caption{}
    \label{fig:1562077004559849}
    \end{subfigure}
    \begin{subfigure}{0.22\textwidth}
    \includegraphics[width=0.8\textwidth, cframe=blue 2pt]{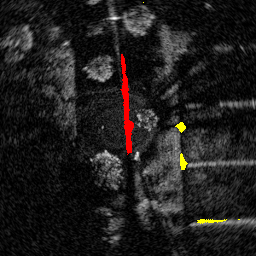}
    \caption{}
    \label{fig:1562076833690941}
    \end{subfigure}
    \begin{subfigure}{0.22\textwidth}
    \includegraphics[width=0.8\textwidth, cframe=cyan 2pt]{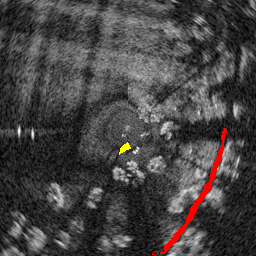}
    \caption{}
    \label{fig:1545308412280131}
    \end{subfigure}
    \begin{subfigure}{0.22\textwidth}
    \includegraphics[width=0.8\textwidth, cframe=cyan 2pt]{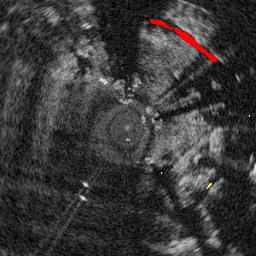}
    \caption{}
    \label{fig:1545308371722058}
    \end{subfigure}
    
    \begin{subfigure}{0.22\textwidth}
    \includegraphics[width=0.8\textwidth, cframe=yellow 2pt]{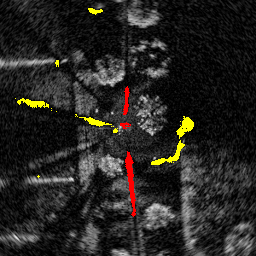}
    \caption{}
    \label{fig:1562077477196799}
    \end{subfigure}
    \begin{subfigure}{0.22\textwidth}
    \includegraphics[width=0.8\textwidth, cframe=yellow 2pt]{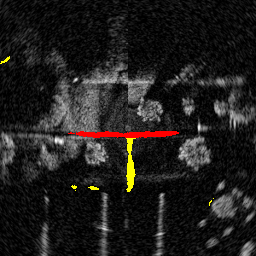}
    \caption{}
    \label{fig:1562077085979532}
    \end{subfigure}
    \begin{subfigure}{0.22\textwidth}
    \includegraphics[width=0.8\textwidth, cframe=orange 2pt]{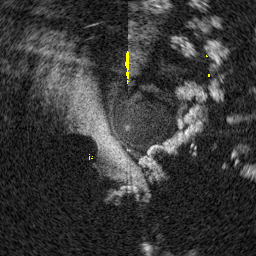}
    \caption{}
    \label{fig:1545308789229384}
    \end{subfigure}
    \begin{subfigure}{0.22\textwidth}
    \includegraphics[width=0.8\textwidth, cframe=orange 2pt]{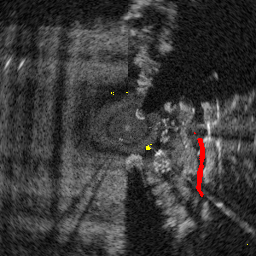}
    \caption{}
    \label{fig:1545308484892422}
    \end{subfigure}
    
    \begin{subfigure}{0.22\textwidth}
    \includegraphics[width=0.8\textwidth, cframe=yellow 2pt]{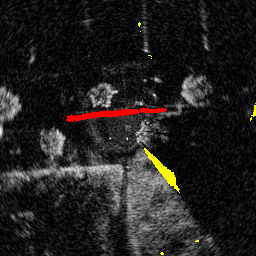}
    \caption{}
    \label{fig:1562077047017197}
    \end{subfigure}
    \begin{subfigure}{0.22\textwidth}
    \includegraphics[width=0.8\textwidth, cframe=yellow 2pt]{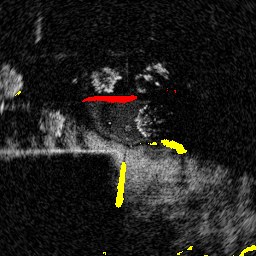}
    \caption{}
    \label{fig:1562077041770728}
    \end{subfigure}
    \begin{subfigure}{0.22\textwidth}
    \includegraphics[width=0.8\textwidth, cframe=orange 2pt]{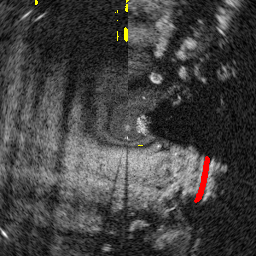}
    \caption{}
    \label{fig:1545308465109444}
    \end{subfigure}
    \begin{subfigure}{0.22\textwidth}
    \includegraphics[width=0.8\textwidth, cframe=orange 2pt]{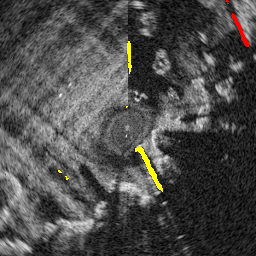}
    \caption{}
    \label{fig:1545308337010979}
    \end{subfigure}
    \caption{U-Net segmentations of test examples in both short and long range configurations (with correct and incorrect predictions in red and yellow respectively). During inference, the dimensions of short and long range scans are \SI{100}{\metre} and \SI{400}{\metre} respectively (c.f.~\cref{sec:application}).
    Scans in the blue and yellow quadrants contain segmentations for the short-range configuration of the radar with yellow containing failure cases. Similarly, the cyan quadrant contains typical long-range examples, while the orange quadrant contains examples of long-range failure cases.
    See~\cref{sec:results:radar} for a discussion of these particular cases.}
    \label{fig:test_results}
    \vspace{-.6cm}
\end{figure*}

Stage two of the curriculum involves fine-tuning the network with the newly generated dataset. 
We then test the network on datasets collected in two unseen locations with the radar in both long and short range configurations.
\cref{fig:test_results} shows both typical segmentations and some radar specific failure cases.

In both short and long range segmentations, the system is able to reliably detect driveable routes with a \SI{360}{\degree} field of view and up to hundreds of metres away.
In \cref{fig:1545308439330558,fig:1545308412280131}, we observe that paths approximately \SI{100}{\metre} away and occluded by trees are accurately segmented in a way that would not be possible using any other sensor modality.
\cref{fig:1562077104209647} shows the network segmenting around pedestrians and \cref{fig:1545308439330558} shows a consistent path detection behind occluding trees.

\cref{fig:1562077477196799,fig:1562077041770728,fig:1545308337010979} show examples where occluded sections of the scan are misclassified as paths.
This problem may be ameliorated by enforcing temporal consistency.
In~\cref{fig:1545308789229384}, the vertical disjoint in the radar scan is misidentified as driveable path.
This artefact arises due to the motion of the radar during scan formation, and can be fixed by motion correction.
Finally, the network understandably doesn't predict through large occlusions, however could be achieved by fitting cubic curves between path segments as in \cite{Suleymanov2018}.

The network correctly classified \SI{98.8}{\percent} of pixels with an achieved IoU score of \SI{39.8}{\percent} when evaluated on 25 hand-labelled unseen examples from the testing set.
Compared with an IoU of \SI{54.1}{\percent} achieved with cameras in \cite{zurn2019self} and considering radar's robustness to weather and illumination shows the feasibility of our method for all-weather scene understanding.

During inference, our U-Net runs at \SI{330}{\hertz} and uses less than \SI{1}{\giga\byte}
of GPU memory when processing \num{256}$\times$\num{256} scans.
We take this to be indicative that a CPU implementation may be feasible for closed-loop autonomy. 
% as the radar update rate is only \SI{4}{\hertz} and new plans would not be required on every new measurement taken by the sensor.

%------------------------------------------------------------------
\section{Conclusions and Future Work}%
\label{sec:conclusions}
%------------------------------------------------------------------
This paper presents a system that identifies permissible driving routes using scanning radar alone.
With a specific focus on the methodology, the system is trained using an audio-leveraged automatic labelling procedure, followed by a curriculum designed to promote generalisation from sparse labelling.
Qualitative results show that the network is capable of generalising effectively to the unseen testing set and to unlabelled areas of the training set. 
Quantitative results demonstrate the feasibility of our methodology for learning robust scene understanding from radar. 

In the future, we plan to retrain and test the system on the all-weather platform described in~\cite{kyberd2019}, as part of closed-loop autonomy.
Specifically, domains for deployment of this platform will be chosen to further explore the generalisibility of the presented system to other grassy environments as well as other driveable surfaces (e.g. asphalt).
The proposed system will also be applied in off-road intelligent transportation contexts~\cite{gadd2020evsav}.

% %------------------------------------------------------------------
% \section*{Acknowledgements}
% %------------------------------------------------------------------

% This project is supported by the Assuring Autonomy International Programme, a partnership between Lloyd’s Register Foundation and the University of York, as well as UK EPSRC Programme Grant EP/M019918/1.
% Additionally, we would like to thank the Groundskeepers and Officers of the University Parks as well as our partners at Navtech Radar.

%------------------------------------------------------------------
\bibliographystyle{IEEEtran}
\bibliography{biblio}

\begin{thebibliography}{10}
\providecommand{\url}[1]{#1}
\csname url@samestyle\endcsname
\providecommand{\newblock}{\relax}
\providecommand{\bibinfo}[2]{#2}
\providecommand{\BIBentrySTDinterwordspacing}{\spaceskip=0pt\relax}
\providecommand{\BIBentryALTinterwordstretchfactor}{4}
\providecommand{\BIBentryALTinterwordspacing}{\spaceskip=\fontdimen2\font plus
\BIBentryALTinterwordstretchfactor\fontdimen3\font minus
  \fontdimen4\font\relax}
\providecommand{\BIBforeignlanguage}[2]{{%
\expandafter\ifx\csname l@#1\endcsname\relax
\typeout{** WARNING: IEEEtran.bst: No hyphenation pattern has been}%
\typeout{** loaded for the language `#1'. Using the pattern for}%
\typeout{** the default language instead.}%
\else
\language=\csname l@#1\endcsname
\fi
#2}}
\providecommand{\BIBdecl}{\relax}
\BIBdecl

\bibitem{cen2018precise}
S.~H. Cen and P.~Newman, ``{Precise ego-motion estimation with millimeter-wave
  radar under diverse and challenging conditions},'' in \emph{IEEE
  International Conference on Robotics and Automation}, 2018.

\bibitem{aldera2019}
R.~Aldera, D.~De~Martini, M.~Gadd, and P.~Newman, ``Fast radar motion
  estimation with a learnt focus of attention using weak supervision,'' in
  \emph{IEEE International Conference on Robotics and Automation}, 2019.

\bibitem{2019ITSC_aldera}
R.~\vspace{0mm}Aldera, D.~De~Martini, M.~Gadd, and P.~Newman, ``{What Could Go
  Wrong? Introspective Radar Odometry in Challenging Environments},'' in
  \emph{{IEEE Intelligent Transportation Systems Conference}}, 2019.

\bibitem{Barnes2019MaskingByMoving}
D.~Barnes, R.~Weston, and I.~Posner, ``{Masking by Moving: Learning
  Distraction-Free Radar Odometry from Pose Information},'' in
  \emph{{C}onference on {R}obot {L}earning ({CoRL})}, 2019.

\bibitem{UnderTheRadarArXiv}
D.~Barnes and I.~Posner, ``Under the radar: Learning to predict robust
  keypoints for odometry estimation and metric localisation in radar,''
  \emph{arXiv preprint arXiv: 2001.10789}, 2020.

\bibitem{KidnappedRadarArXiv}
{\cb{S}}.~S\u{a}ftescu, M.~Gadd, D.~De~Martini, D.~Barnes, and P.~Newman,
  ``{Kidnapped Radar: Topological Radar Localisation using
  Rotationally-Invariant Metric Learning},'' \emph{arXiv preprint arXiv:
  2001.09438}, 2020.

\bibitem{gadd2020lookaroundyou}
M.~Gadd, D.~De~Martini, and P.~Newman, ``{Look Around You: Sequence-based Radar
  Place Recognition with Learned Rotational Invariance},'' in \emph{IEEE/ION
  Position, Location and Navigation Symposium}, 2020.

\bibitem{tang2020rsl}
T.~Y. Tang, D.~De~Martini, D.~Barnes, and P.~Newman, ``{RSL-Net: Localising in
  Satellite Images From a Radar on the Ground},'' \emph{arXiv preprint
  arXiv:2001.03233}, 2020.

\bibitem{williams2019listening}
D.~Williams, D.~De~Martini, L.~Marchegiani, and P.~Newman, ``{Listening closely
  to see far away: Radar-based terrain classification from auditory signals},''
  in \emph{{International Conference on Digital Image and Signal Processing
  (DISP)}}, 2019.

\bibitem{weston2019probably}
R.~Weston, S.~Cen, P.~Newman, and I.~Posner, ``{Probably Unknown: Deep Inverse
  Sensor Modelling Radar},'' in \emph{2019 International Conference on Robotics
  and Automation}, 2019.

\bibitem{kaul2020rssnet}
P.~Kaul, D.~De~Martini, M.~Gadd, and P.~Newman, ``{RSS-Net: Weakly-Supervised
  Multi-Class Semantic Segmentation with FMCW Radar},'' in \emph{IEEE
  Intelligent Vehicles Symposium}, 2020.

\bibitem{robo_radar}
E.~J. M.~Adams, J.~Mullane and B.~Vo, \emph{{Robot Navigation and Mapping with
  Radar}}.\hskip 1em plus 0.5em minus 0.4em\relax Artech House, 2012.

\bibitem{cordts2016cityscapes}
M.~Cordts, M.~Omran, S.~Ramos, T.~Rehfeld, M.~Enzweiler, R.~Benenson,
  U.~Franke, S.~Roth, and B.~Schiele, ``The cityscapes dataset for semantic
  urban scene understanding,'' in \emph{IEEE conference on computer vision and
  pattern recognition}, 2016.

\bibitem{barnes16}
D.~Barnes, W.~P. Maddern, and I.~Posner, ``Find your own way: Weakly-supervised
  segmentation of path proposals for urban autonomy,'' \emph{CoRR}, vol.
  abs/1610.01238, 2016.

\bibitem{blas2008fast}
M.~R. Blas, M.~Agrawal, A.~Sundaresan, and K.~Konolige, ``Fast color/texture
  segmentation for outdoor robots,'' in \emph{IEEE/RSJ International Conference
  on Intelligent Robots and Systems}, 2008.

\bibitem{jansen2005colour}
P.~Jansen, W.~van~der Mark, J.~C. van~den Heuvel, and F.~C. Groen, \emph{Colour
  based off-road environment and terrain type classification}.\hskip 1em plus
  0.5em minus 0.4em\relax Piscataway, NJ: IEEE, 2005.

\bibitem{kragh2015object}
M.~Kragh, R.~N. J{\o}rgensen, and H.~Pedersen, ``Object detection and terrain
  classification in agricultural fields using 3d lidar data,'' in
  \emph{International conference on computer vision systems}, 2015.

\bibitem{Valada2018}
A.~Valada, L.~Spinello, and W.~Burgard, \emph{{Deep Feature Learning for
  Acoustics-Based Terrain Classification BT - Robotics Research: Volume 2}},
  A.~Bicchi and W.~Burgard, Eds.\hskip 1em plus 0.5em minus 0.4em\relax
  Springer, 2018.

\bibitem{mielle2019comparative}
M.~Mielle, M.~Magnusson, and A.~J. Lilienthal, ``{A comparative analysis of
  radar and lidar sensing for localization and mapping},'' in \emph{European
  Conference on Mobile Robotics (ECMR)}, 2019.

\bibitem{heuer2014detection}
M.~Heuer, A.~Al-Hamadi, A.~Rain, and M.-M. Meinecke, ``Detection and tracking
  approach using an automotive radar to increase active pedestrian safety,'' in
  \emph{IEEE Intelligent Vehicles Symposium}, 2014.

\bibitem{brooker2007seeing}
G.~Brooker, R.~Hennessey, C.~Lobsey, M.~Bishop, and E.~Widzyk-Capehart,
  ``Seeing through dust and water vapor: Millimeter wave radar sensors for
  mining applications,'' \emph{Journal of Field Robotics}, vol.~24, no.~7,
  2007.

\bibitem{foessel1999short}
A.~Foessel, S.~Chheda, and D.~Apostolopoulos, \emph{Short-range millimeter-wave
  radar perception in a polar environment}.\hskip 1em plus 0.5em minus
  0.4em\relax Carnegie Mellon University, 1999.

\bibitem{reina2011radar}
G.~Reina, J.~Underwood, G.~Brooker, and H.~Durrant-Whyte, ``Radar-based
  perception for autonomous outdoor vehicles,'' \emph{Journal of Field
  Robotics}, vol.~28, no.~6, 2011.

\bibitem{zurn2019self}
J.~Z{\"u}rn, W.~Burgard, and A.~Valada, ``Self-supervised visual terrain
  classification from unsupervised acoustic feature learning,'' \emph{arXiv
  preprint arXiv:1912.03227}, 2019.

\bibitem{Marchegiani2018}
L.~Marchegiani and P.~Newman, ``Listening for sirens: Locating and classifying
  acoustic alarms in city scenes,'' \emph{ArXiv}, vol. abs/1810.04989, 2018.

\bibitem{curriculum_learning}
Y.~Bengio, J.~Louradour, R.~Collobert, and J.~Weston, ``Curriculum learning,''
  \emph{Journal of the American Podiatry Association}, vol.~60, 2009.

\bibitem{labelrefinery}
H.~Bagherinezhad, M.~Horton, M.~Rastegari, and A.~Farhadi, ``Label refinery:
  Improving imagenet classification through label progression,'' \emph{CoRR},
  vol. abs/1805.02641, 2018.

\bibitem{ronneberger2015u}
O.~Ronneberger, P.~Fischer, and T.~Brox, ``U-net: Convolutional networks for
  biomedical image segmentation,'' in \emph{International Conference on Medical
  image computing and computer-assisted intervention}, 2015.

\bibitem{Suleymanov2018}
T.~Suleymanov, P.~Amayo, and P.~Newman, ``Inferring road boundaries through and
  despite traffic,'' \emph{2018 21st International Conference on Intelligent
  Transportation Systems}, 2018.

\bibitem{kyberd2019}
S.~Kyberd, J.~Attias, P.~Get, P.~Murcutt, C.~Prahacs, M.~Towlson, S.~Venn,
  A.~Vasconcelos, M.~Gadd, D.~De~Martini, and P.~Newman, ``{The Hulk: Design
  and Development of a Weather-proof Vehicle for Long-term Autonomy in Outdoor
  Environments},'' in \emph{{International Conference on Field and Service
  Robotics}}, 2019.

\bibitem{gadd2020evsav}
M.~Gadd, D.~De~Martini, L.~Marchegiani, L.~Kunze, and P.~Newman,
  ``{Sense-Assess-eXplain (SAX): Building Trust in Autonomous Vehicles in
  Challenging Real-World Driving Scenarios},'' in \emph{IEEE Intelligent
  Vehicles Symposium, Workshop on Ensuring and Validating Safety for Automated
  Vehicles (EVSAV)}, 2020.

\end{thebibliography}
%------------------------------------------------------------------

\end{document}